\documentclass{article}

\usepackage{times}
\usepackage{graphicx} 
\usepackage{subfigure} 

\usepackage{natbib}

\usepackage{hyperref}       
\usepackage{url}            
\usepackage{booktabs}       
\usepackage{amsmath,amsfonts}       
\usepackage{nicefrac}       
\usepackage{microtype}      
\usepackage{graphicx}
\usepackage{xspace,color}
\usepackage{rotating}

\usepackage{tikz}
\usepackage{tikzscale}
\usetikzlibrary{positioning}
\usetikzlibrary{shapes.misc}
\usetikzlibrary{arrows.meta}

\definecolor{inriablue}{RGB}{0, 96, 215}

\usepackage{algorithm}
\usepackage{algorithmic}

\usepackage{hyperref}

\usepackage[accepted]{icml2017}

\definecolor{darkgreen}{RGB}{0, 140, 0}

\def \ie {\emph{i.e.}\xspace}
\def \eg {\emph{e.g.}\xspace}

\def \Vcal {\mathcal V}

\def \Ocal {\mathcal O}

\def \Vh {{\mathcal V}_\mathrm{h}}
\def \kh {k_\mathrm{h}}

\icmltitlerunning{Efficient softmax approximation for GPUs}

\begin{document} 

\twocolumn[
\icmltitle{Efficient softmax approximation for GPUs}




\begin{icmlauthorlist}
\icmlauthor{\'Edouard Grave}{all}
\icmlauthor{Armand Joulin}{all}
\icmlauthor{Moustapha Ciss\'e}{all}
\icmlauthor{David Grangier}{all}
\icmlauthor{Herv\'e J\'egou}{all}
\end{icmlauthorlist}

\icmlaffiliation{all}{Facebook AI Research}

\icmlcorrespondingauthor{\'Edouard Grave}{egrave@fb.com}


\vskip 0.3in
]



\printAffiliationsAndNotice{}  

\begin{abstract} We propose an approximate strategy to efficiently train neural
network based language models over very large vocabularies.
Our approach, called adaptive softmax, circumvents the linear dependency on the vocabulary size by
exploiting the unbalanced word distribution to form clusters that explicitly
minimize the expectation of computation time.  Our approach further
reduces the computational time by exploiting the specificities of modern
architectures and matrix-matrix vector operations, making it
particularly suited for graphical processing units. Our experiments carried out 
on standard benchmarks, such as EuroParl and One Billion Word, show that our 
approach brings a large gain in efficiency over standard approximations while
achieving an accuracy close to that of the full softmax.
The code of our method is available at {\small\url{https://github.com/facebookresearch/adaptive-softmax}}.
\end{abstract}

\section{Introduction}
\label{sec:introduction}

This paper considers strategies to learn parametric models for language
modeling with very large vocabularies. This problem is key to natural language
processing, with applications in machine
translation~\citep{schwenk2012large,sutskever2014sequence,vaswani2013decoding}
or automatic speech
recognition~\citep{graves2013speech,hinton2012deep}.  In
particular, Neural Network Language Models (NNLMs) have received a renewed
interest in recent years, by achieving state of the art performance on standard
benchmarks~\citep{jozefowicz2016exploring,mikolov2010recurrent}.  These
approaches are more computationally intensive but generalize better than traditional
non-parametric models~\citep{bahl1983maximum,kneser1995improved}.

Statistical language models assign a probability to words given their
history~\citep{bahl1983maximum}. They are evaluated by objective criteria such as
perplexity (ppl), which directly measures the ability of the system to determine 
proper probabilities for all the words. This potentially makes parametric models
prohibitively slow to train on corpora with very large vocabulary. For instance, the vocabulary of 
the One Billion Word benchmark~\citep{chelba2013one} contains around $800$K words.
In standard NNLMs, such as feedforward networks~\citep{bengio2003neural} or
recurrent networks~\citep{mikolov2010recurrent}, computing this probability over
the whole vocabulary is the bottleneck.  Many solutions have been proposed to
reduce the complexity of this expensive
step~\citep{bengio2003quick,goodman2001classes,gutmann2010noise}. We distinguish
(i) the methods that consider the original distribution and aim at providing
approximations of the probabilities, or
of a subset of them~\citep{bengio2003quick,ji2015blackout}, from (ii) the approaches that compute exact probabilities
for an approximate model yielding a lower computational time, such as the
popular hierarchical
softmax~\citep{goodman2001classes,mnih2009scalable,morin2005hierarchical}. 

Our approach, called adaptive softmax, belongs to the second category. More specifically, it is inspired by
the hierarchical softmax and its subsequent variants. 
In contrast to previous works and motivated by the trend that GPUs are comparatively more and
more performant than CPUs, our design is oriented towards efficient processing on GPUs. 
In this context, our paper makes the following points:
\begin{itemize} 
\item We define a strategy to produce an approximate
hierarchical model. It departs from previous ones in that it explicitly takes
into account the computation time of matrix-matrix multiplications on modern
architectures, which is not trivially linear in the dimensions of the matrices.
\item We conduct an empirical analysis of this model on recent GPUs.
This leads us to define a realistic computation time model that is incorporated in
the proposed optimization; 
\item Our approach provides a
significant acceleration factor compared to the regular softmax, i.e., $2\times$ to $10\times$ speed-ups. 
Equivalently we improve the accuracy under computational constraints.
Importantly, on the largest corpus, this higher efficiency empirically comes at 
no cost in accuracy for a given amount of training data, in contrast to concurrent approaches improving the efficiency.   
\end{itemize}

This paper is organized as follows. Section~\ref{sec:relatedwork} briefly
reviews the related work and Section~\ref{sec:preliminaries} provides some
background on the language modeling task that we consider.
Section~\ref{sec:approach} describes our proposal, which is subsequently
evaluated in Section~\ref{sec:experiments} on typical benchmarks of the
language modeling literature, including Text8, Europarl and One
Billion Word datasets.

\section{Related work}
\label{sec:relatedwork}

Many methods have been proposed to approximate the softmax
efficiently~\citep{bengio2003quick,goodman2001classes,gutmann2010noise,morin2005hierarchical}.
We briefly describe the most popular ones below and point the reader to \citet{chen2015strategies}
for a comparative study. For the sake of completeness, we refer the reader to other strategies that can speed-up the
training of language models in complementary
manners~\citep{mikolov2011strategies}.

\paragraph{Loss function approximation.}

The \emph{Hierarchical Softmax} (HSM) is an approximation of the softmax function
introduced by \citet{goodman2001classes}.  This approach is generally used
with a two-level tree~\citep{goodman2001classes,mikolov2011extensions} but has
also been extended to deeper hierarchies~\citep{morin2005hierarchical,mnih2009scalable}.
In general, the hierarchy structure is built on word
similarities~\citep{brown1992class,le2011structured,mikolov2013efficient} or
frequency binning~\citep{mikolov2011extensions}.
In particular, \citet{mikolov2013efficient} proposes an optimal
hierarchy by constructing a Huffman coding based on frequency. 
However this coding scheme does not take into account the theoretical
complexity reduction offered by matrix-matrix multiplication and distributed
computation, in particular with modern GPUs.

Similar to our work, \citet{zweig2013speed} constructs their hierarchy
in order to explicitly reduce the computational complexity. They also solve
the assignment problem with dynamic programming.  However, they only
consider hierarchies where words are kept in the leaves of the tree, leading
to a significant drop of performance (reported to be around $5-10\%$), forcing
them to also optimize for word similarity.  In our case, allowing classes to be
stored in the internal node of the tree leads to almost no drop of performance.
Also, they assume a linear computational time for the vector-matrix operation which
significantly limits the use of their approach on distributed system such as GPU. 

The idea of keeping a short-list of the most frequent words has been explored
before~\citep{le2011structured,schwenk2007continuous}.  In particular,
\citet{le2011structured} combines a short-list with a hierachical softmax based on
word representation. In contrast, the word hierarchy that we introduce in
Section~\ref{sec:approach} explicitly aims at reducing the complexity.

Our work also shares similarities with the \emph{d-softmax}
introduced by \citet{chen2015strategies}. They assign capacity to
words according to their frequency to speed up the training. Less frequent
words have smaller classifiers than frequent ones. Unlike our method, their
formulation requires accessing the whole vocabulary to evaluate the probability of a word.

\paragraph{Sampling based approximation.} 
Sampling based approaches have been successfully applied to approximate the
softmax function over large dictionaries in different domains, such as language
modeling~\citep{jozefowicz2016exploring}, machine translation~\citep{jean2015}
and computer vision~\citep{joulin2015learning}.  In particular, importance
sampling~\citep{bengio2008adaptive,bengio2003quick} selects a subset of
negative targets to approximate the softmax normalization.  Different schemes
have been proposed for sampling, such as the unigram and bigram
distribution~\citep{bengio2003quick} or more recently, a power-raised
distribution of the unigram~\citep{ji2015blackout,mikolov2013efficient}.  While
this approach often leads to significant speed-up at train time, it still requires
to evaluate the full softmax at test time.

\paragraph{Self-normalized approaches.}

Self-normalized approaches aim at learning naturally normalized classifier, to
avoid computing the softmax normalization. Popular methods are Noise
Contrastive Estimation~\citep{gutmann2010noise,mnih2012fast,vaswani2013decoding}
or a penalization on the normalization function~\citep{andreas2014and,devlin2014fast}.  Noise Contrastive
Estimation~\citep{gutmann2010noise} replaces the softmax by a binary classifier
distinguishing the original distribution form a noisy one.  While the original
formulation still requires to compute the softmax normalization, \citet{mnih2012fast}
shows that good performance can be achieved even without it. 

Finally, \citet{vincent2015efficient} have also proposed an
efficient way to train model with high dimensional output space.  Their approach
is exact and leads to a promising speed-up but it cannot be directly applied to
the softmax function, limiting its potential application to language modeling.

\section{Preliminaries on language modeling}
\label{sec:preliminaries}

The goal of language modeling is to learn a probability distribution over a
sequence of words from a given dictionary $\Vcal$.
The joint distribution is defined as a product of
conditional distribution of tokens given their past~\citep{bahl1983maximum}.
More precisely, the probability of a sequence of $T$ words $w_1,\dots,w_T \in \Vcal^T$ is given as
\begin{align}
P(w_1,\dots,w_T) = \prod_{t=1}^T P(w_t~|~w_{t-1},\dots,w_1).
\end{align}
This problem is traditionally addressed with non-parameteric models based on
counting statistics~\citep{goodman2001bit}. In particular, smoothed N-gram
models~\citep{bahl1983maximum,katz1987estimation,kneser1995improved}
achieve good performance in practice~\citep{mikolov2011empirical},
especially when they are associated with cache models~\citep{kuhn1990cache}.
More recently, parametric models based on neural networks have gained popularity
for language modeling~\citep{bengio2003neural,jozefowicz2016exploring,mikolov2010recurrent}.
They are mostly either feedforward networks~\citep{bengio2003neural} or recurrent networks~\citep{mikolov2010recurrent}.

\subsection{Feedforward network.}
In a standard feedforward network for language modeling, we fix a window of
length $N$ and predict the next words according to the words appearing in this
window.
In the simplest case, this probability is represented by a 2-layer neural
network acting on an input $x_t \in \Vcal^N$, defined as the
concatenation of the one-hot representation of the $N$ previous words, 
$w_{t-N+1},\dots, w_{t}$. The state $h_t$ of the hidden layer 
and subsequently the vector of scores $y_t$
associated with the next token $w_{t+1}$ are computed as 
\begin{align}
h_t & = \sigma( A P x_t ), \\ 
y_t & = f(B h_t),
\end{align}
where $\sigma$ is a non linearity, e.g., the pointwise sigmoid function $\sigma(z) = 1 / (1 + \exp(-z) )$, and $f$
is the softmax function discussed in section~\ref{sec:softmax}.
This model is parameterized by the weight matrices $P$, $A$ and $B$ and is routinely 
learned with an optimization scheme such as stochastic gradient descent
or Adagrad~\citep{duchi2011adaptive}.

\subsection{Recurrent network.} A Recurrent network~\citep{elman1990finding} extends  
a feedforward network in that the current state of the hidden layer also depends on its previous state. 
The hidden state $h_t$ is updated according to the equation
$$
h_t = \sigma( A w_t +R h_{t-1}),
$$
where $R$ is a weight matrix and $x_t$ is the one-hot representation of the
current word $w_t$.  Computing the exact gradient for this model is challenging
but it is possible to compute an efficient and stable approximation of it,
using a truncated back-propagation through
time~\citep{werbos1990backpropagation, williams1990efficient} and norm
clipping~\citep{mikolov2010recurrent}.

Since the model introduced by \citet{elman1990finding}, many
extensions have been proposed, such as Longer Short Term Memory
(LSTM)~\citep{hochreiter1997long}, Gated recurrent
units~\citep{chung2014empirical} or structurally constrained
network~\citep{mikolov2014learning}.
These models have been successfully used in the context of language
modeling~\citep{jozefowicz2016exploring,mikolov2010recurrent,mikolov2012context}.
In this work, we focus on the standard word level LSTM architecture
since it has obtained state of the art performance on the challenging
One Billion Word Benchmark~\citep{jozefowicz2016exploring}.

\subsection{Class-based hierarchical softmax.}
\label{sec:softmax}
In neural language modeling, predicting the probability of the next word requires computing scores for every word in the vocabulary and
 to normalize them to form a probability distribution.
This is typically achieved by applying a softmax function to the unnormalized score $z_w$ associated with each word $w$, where the softmax function is defined as
\begin{align} f(z_w) =
\frac{\exp(z_w)}{\sum_{w' \in \Vcal} \exp(z_{w'})}.
\label{eq:softmax}
\end{align}
For a vocabulary comprising $k=|\Vcal|$ words, this function requires $\Ocal(k)$ operations once  the scores are computed.
In the case of neural networks, the overall complexity is $\Ocal(d k)$, where $d$ is the size of the last hidden layer.
When the vocabulary is large, this step is computationally expensive and often dominates the computation of the whole 
model~\citep{jozefowicz2016exploring,mikolov2014learning}, as discussed in introduction and related work.
A simple approach~\citep{goodman2001classes} to reduce this computational cost is to assign each word $w$ of the vocabulary to a unique class $\mathcal{C}(w)$ and
to factorize the probability distribution over words as
\begin{equation*}
p(w_t \ | \ h_t) = p_1(\mathcal{C}(w_t) \ | \ h_t) \times p_2(w_t \ | \ \mathcal{C}(w_t), \ h_t),
\end{equation*}
where $p_1$ and $p_2$ are obtained using the softmax function (Eq.~\ref{eq:softmax}).
If each class contains $\sqrt{k}$ words, the computational cost is reduced from $\Ocal(d k)$ to $\Ocal(d \sqrt{k})$.

\section{Our approach: the adaptive softmax}
\label{sec:approach}

In this section, we propose the adaptive softmax, a simple speedup technique for the computation of probability distributions over words.
The adaptive softmax is inspired by the class-based hierarchical softmax, where the word classes are built to minimize the computation time.
Our method is designed to be efficient for GPUs, which are commonly used to train neural networks.
For the sake of clarity, we first present the intuition behind our method in the simple case 
where we simply split our dictionary in two distinct clusters, before analyzing a more general case.

\begin{figure}[t]
\includegraphics[width=0.99 \linewidth]{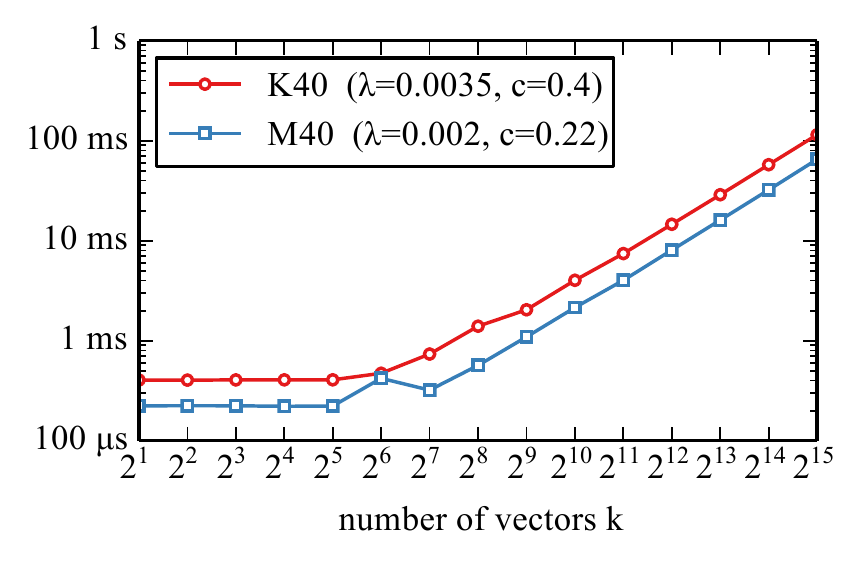}
\caption{GPU timings for multiplying two matrices.
We consider matrices of size $2560\times 2048$ and $2048\times k$ representing hidden states and word representations.
We report average timings over 1000 measures as a function of $k$ (number of words). 
\label{fig:cost_mm_gpu}}
\end{figure}

\subsection{Computation time model of matrix-multiplication}

The bottleneck of the model described in the previous section is the matrix multiplication between the matrix representing the hidden states (of size $B \times d$, where $B$ denotes the batch size), and the matrix of word representations, of size $d\times k$.
For a fixed size $d$ of the hidden layer, we denote by $g(k, B)$ the computation time of this multiplication (using an efficient implementation such as cuBLAS), and simplify the notation wherever some parameters are fixed.
Figure~\ref{fig:cost_mm_gpu} reports empirical timings as a function of $k$ for typical parameters of $B$ and $d$ for two GPU models, namely K40 and M40.
We observe that the computation time $g(k)$ is constant for low values of $k$, until a certain inflection point $k_0\approx 50$, and then becomes affine for values $k > k_0$.
This suggests a computational model of the form 
\begin{align}
g(k) & = \max (c + \lambda k_0 , c + \lambda k) \\
& = c_\mathrm{m} + \max \big[0,\lambda (k-k_0)\big]. 
\end{align}
Empirically, $c_\mathrm{m}=0.40$\,ms on a K40 and 0.22\,ms on a M40.
We observe the same behavior when measuring the timings as a function of the batch size $B$, \ie, it is inefficient to matrix-multiplication when one of the dimensions is small.
This observation suggests that hierarchical organizations of words with a low number of children per node, such as binary Huffman codes, are highly suboptimal. 
Similarly, clusters comprising only rare words have a low probabilty $p$ and a shrinking batch size of $p\ B$, which also lead to iniffient matrix-multiplication.
In the following, we propose to use the following model of computation time for matrix-multiplication
\begin{equation}
g(k, B) = \max(c + \lambda k_0 B_0, c + \lambda k B).
\label{equ:costmodel}
\end{equation}
While this is a very crude model of computation, it allows to explain empirical observations well.

\subsection{Intuition: the two-clusters case}

In natural languages, the distribution of the words notoriously follows a Zipf
law~\citep{zipf1949human}.  Most of the probability mass 
is covered by a small fraction of the dictionary, \eg, 
$87\%$ of the document is covered by only $20\%$ of the vocabulary in the Penn TreeBank.
Similar to the frequency binning hierarchical
softmax~\citep{mikolov2011extensions}, this information can be exploited
to reduce the computation time. 

A simple strategy to reduce the overall computation time is to partition the dictionary $\Vcal$ into two clusters as $\Vcal_\mathrm{h}$ and $\Vcal_\mathrm{t}$, where $\Vcal_\mathrm{h}$ denotes
the \emph{head} of the distribution consisting of the most frequent words, and where $\Vcal_\mathrm{t}$ is the \emph{tail} associated with a large number of rare words. 
The classifier frequently accesses the head, which motivates the fact that it should be computed efficiently. 
In contrast, the tail occurs less frequently and the corresponding computation can be slower. 
This suggests defining clusters with unbalanced cardinalities $|\Vcal_\mathrm{h}| \ll |\Vcal_\mathrm{t}|$ and 
probabilities $P(\Vcal_\mathrm{h}) \gg P(\Vcal_\mathrm{t})$, where $P({\mathcal A}) = \sum_{w \in {\mathcal A}} p_i$ is the probability of a word to occur in the set $\Vcal_i$. 
For instance, one may define the head would only contain
$20\%$ of the vocabulary (covering for $87\%$ on PennTree Bank). 
These two clusters can be organized in two different ways: either they are
both leaves of a 2-level tree~\citep{mikolov2011extensions}, or the head cluster
is kept as a \emph{short-list} in the root node~\citep{le2011structured}.

\paragraph{Compromising between efficiency and accuracy.} We observe
empirically that putting all the clusters in the leaves of the tree leads to a
significant drop of performance~\citep[around~$5-10\%$ performance
drop,][]{mikolov2011extensions,zweig2013speed}.  The reason is that the
probability of every word $w$ belonging to a cluster $c$ is multiplied by the
probability of its class, i.e., it is equal to $P(c~|~h)P(w~|~c,~h)$, while attaching a frequent word directly to the root associates it directly to the probability $P(w~|~h)$ making its inference sharper.
For this reason, unless there is a significant difference in computation time, we
favor using a short-list, over the standard 2-level hierarchical softmax.

\begin{figure}[t]
\centering
\includegraphics[scale=0.9]{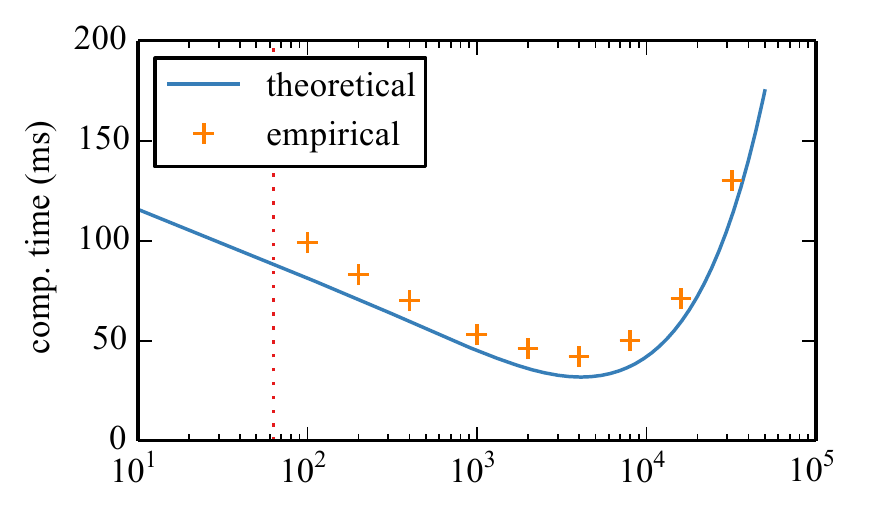}
\caption{Computational time for the two-clusters adaptive softmax on the Bulgarian Europarl data, as a function of the size of the head cluster $k_h$.
We report values predicted by our model (\texttt{theoretical}) as well as measured on a K40 (\texttt{empirical}).
Even with this very simple hierarchy, we observe more than $5 \times$ speedup over the full softmax.
The red dotted line indicates the value of the parameter $k_h$ such that both clusters have equal probability: $p_h = p_t = 0.5$.
\label{fig:cost_europarl_bg}}
\end{figure}

\paragraph{Minimizing the computation time.}
Given a vocabulary of $k$ words, we are looking for the number $k_\mathrm{h}=|\Vcal_\mathrm{h}|$ of words from the head of the distribution to be assigned to the first cluster.
These words will cover for $p_\mathrm{h}$ of the distribution. The tail cluster will then contain the rest of the vocabulary, made of $k_\mathrm{t} = k - k_\mathrm{h}$ words and covering for $p_\mathrm{t} = 1 - p_\mathrm{h}$ of the overall distribution.
The computation time corresponding to the matrix multiplication of the root is equal to $g(k_\mathrm{h} + 1, B)$, while the computation time for the tail of the distribution is equal to $g(k_\mathrm{t}, p_\mathrm{t} B)$, where $B$ is the batch size.
We thus obtain the overall computation time
$$
C = g(k_\mathrm{h} + 1, B) + g(k_\mathrm{t}, p_\mathrm{t} B).
$$
We can then find the size of the head cluster $k_\mathrm{h}$ which minimizes the computation time $C$.
We plot the value of $C$ as a function of $k_\mathrm{h}$ in Figure~\ref{fig:cost_europarl_bg}, for the word distribution of the Bulgarian Europarl dataset.
We observe that the optimal splitting between head and tail gives a $5 \times$ speedup over the full softmax.
Another important observation is the fact that the optimal size of the head cluster does not correspond to two clusters with equal probability.


\paragraph{Adapting the classifier capacity for each cluster.} 
Each cluster is accessed independently of each other, they thus do not need to
have the same capacity.  Frequent words
need high capacity to be predicted correctly.  
In contrast, rare words cannot be learned very well, since we only see them
a few times. It would then be wasteful to associate them with
high capacity. Like in \citet{chen2015strategies}, we exploit this
observation to further reduce the computational time of our classifier.
Unlike \citet{chen2015strategies}, we share the state of hidden layer across clusters
and simply reduce the input size of the classifiers by applying a projection
matrix.  Typically, the projection matrix for the tail cluster reduces the size
from $d$ to $d_\mathrm{t}=d/4$.

\subsection{General case}
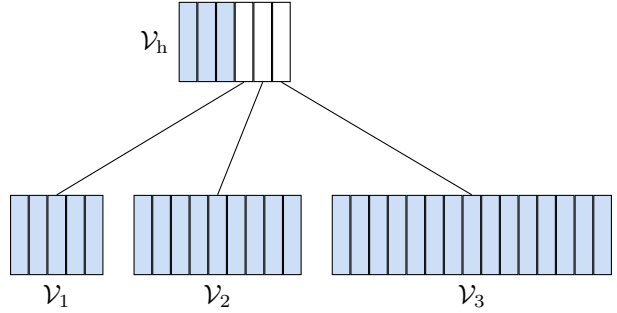
\begin{figure}
  \centering
  \begin{tikzpicture}[node distance=0]
    \tikzstyle{hidden}=[draw=black,minimum height=30pt]
    \tikzstyle{cell}=[draw=black,fill=inriablue!20,minimum height=30pt]
    \tikzstyle{arrow}=[-{Latex[scale=1.0]},draw=black]

    \node (c1) [cell] {};
    \node (c2) [right=of c1,cell] {};
    \node (c3) [right=of c2,cell] {};
    \node (c4) [right=of c3,hidden] {};
    \node (c5) [right=of c4,hidden] {};
    \node (c6) [right=of c5,hidden] {};

    \node (c1w1) [below left=1.5cm and 1cm of c1,cell] {};
    \node (c1w2) [left=of c1w1,cell] {};
    \node (c1w3) [left=of c1w2,cell] {};
    \node (c1w4) [left=of c1w3,cell] {};
    \node (c1w5) [left=of c1w4,cell] {};

    \node (c4w1) [right=0.4cm of c1w1,cell] {};
    \node (c4w2) [right=of c4w1,cell] {};
    \node (c4w3) [right=of c4w2,cell] {};
    \node (c4w4) [right=of c4w3,cell] {};
    \node (c4w5) [right=of c4w4,cell] {};
    \node (c4w6) [right=of c4w5,cell] {};
    \node (c4w7) [right=of c4w6,cell] {};
    \node (c4w8) [right=of c4w7,cell] {};
    \node (c4w9) [right=of c4w8,cell] {};

    \node (c5w1) [right=0.4cm of c4w9,cell] {};
    \node (c5w2) [right=of c5w1,cell] {};
    \node (c5w3) [right=of c5w2,cell] {};
    \node (c5w4) [right=of c5w3,cell] {};
    \node (c5w5) [right=of c5w4,cell] {};
    \node (c5w6) [right=of c5w5,cell] {};
    \node (c5w7) [right=of c5w6,cell] {};
    \node (c5w8) [right=of c5w7,cell] {};
    \node (c5w9) [right=of c5w8,cell] {};
    \node (c5w10) [right=of c5w9,cell] {};
    \node (c5w11) [right=of c5w10,cell] {};
    \node (c5w12) [right=of c5w11,cell] {};
    \node (c5w13) [right=of c5w12,cell] {};
    \node (c5w14) [right=of c5w13,cell] {};
    \node (c5w15) [right=of c5w14,cell] {};

    \node (aaa) [left=of c1] {$\mathcal{V}_\mathrm{h}$};
    \node (bbb) [below=of c1w3] {$\mathcal{V}_\mathrm{1}$};
    \node (ccc) [below=of c4w5] {$\mathcal{V}_\mathrm{2}$};
    \node (ddd) [below=of c5w8] {$\mathcal{V}_\mathrm{3}$};

    \draw (c6.south) -- (c5w8.north);
    \draw (c5.south) -- (c4w5.north);
    \draw (c4.south) -- (c1w3.north);

  \end{tikzpicture}
\caption{Our hierarchical model is organized as (i) a first level that includes both the most frequent words and  vectors representing clusters, and (ii) clusters on the second level that are associated with rare words, the largest ones being associated with the less frequent words.
The sizes are determined so as to minimize our computational model on GPU. 
\label{fig:hierarchy}}
\end{figure}

Let us now consider the more general case where the dictionary is partitioned as $\Vcal = \Vh \cup \Vcal_1 \dots \Vcal_J$, $\Vcal_i \cap \Vcal_j = \emptyset$ if $i \neq j$. We consider the hierarchical model depicted in Figure~\ref{fig:hierarchy}, where the sub-dictionary $\Vh$ is accessed at the first level, and the others in the second level. We now investigate the computational cost $C$ of the forward (equivalently, backward) pass of this approximate softmax layer. For the time being, we fix the batch size $B$ and the dimensionality $d$ of the hidden layer, in order to analyze the computation time as a function of the sub-dictionary sizes and probabilities. We denote by $p_i = \sum_{w \in {\mathcal \Vcal_i}} p(w)$ the probability $P(w \in \Vcal_i)$ and $k_i=|\Vcal_i|$ the cardinality of each cluster. 

The expected computational cost $C$ is decomposed as $C = C_\mathrm{h} + \sum_i C_i$, where 
$$ C_\mathrm{h} = g(J+\kh,\ B) $$
and
$$ \forall i,\ C_i = g(k_i,\ p_i\ B), $$
leading to 
\begin{equation}
C = g(J+\kh,\ B) + \sum_{i} g(k_i,\ p_i B).
\end{equation}

We add the constraint $k B \geq k_0 B_0$ to ensure that there is no penalty induced by the constant part of the computational model of Equation~\ref{equ:costmodel}, the previous equation simplifies as 
\begin{align}
C & = c + \lambda (J+\kh)B + \sum_i \left(c + \lambda k_i p_i B\right) \\
  & = (J+1)c + \lambda B \big[J+\kh + \sum_i  p_i \, k_i \big].
\label{equ:totalcost}
\end{align} 

Let us discuss this equation, by first considering that the cardinalities of the sub-vocabularies are fixed. The right-most term is the only one that depends on the word probabilities. For two distinct clusters $\Vcal_i$ and $\Vcal_j$, we can re-write $p_j k_j$ as $(p_{i+j} - p_i) k_j$, where $p_{i+j}=p_i+p_j$, so that
\begin{equation}
p_i k_i + p_j k_j = p_i (k_i - k_j) + p_{i+j} k_j. 
\end{equation}
Without loss of generality, we assume that $k_i > k_j$. 
The quantities $p_{i+j}$, $k_i$ and $k_j$ being fixed, the second term of the right-hand side of this equation is constant, and the best strategy is trivially to minimize the probability of the largest cluster $\Vcal_i$. In other terms, an optimal solution for Equation~\ref{equ:totalcost} requires that the most frequent words are assigned to the smallest cluster. This remark is true for any tuple $(i,j)$, and we easily see that this point also holds for the head cluster. As a consequence, for a fixed number of clusters of given sizes, the best strategy is to assign the words by decreasing probabilities to clusters of increasing size. Note, this analysis remains valid as long as the $g$ is monotonically increasing in $k$. 

\paragraph{Determining $k_i$ with $J$ fixed: dynamic programming.} We now assume that the number of clusters is fixed. Following our analysis above, the optimization solely depends on the cardinalities $k_i$ for all clusters, which perfectly determines how to split the list of words ordered by frequency. We solve this problem by dynamic programming. 

\begin{figure}[t]
\centering
\includegraphics[scale=0.9]{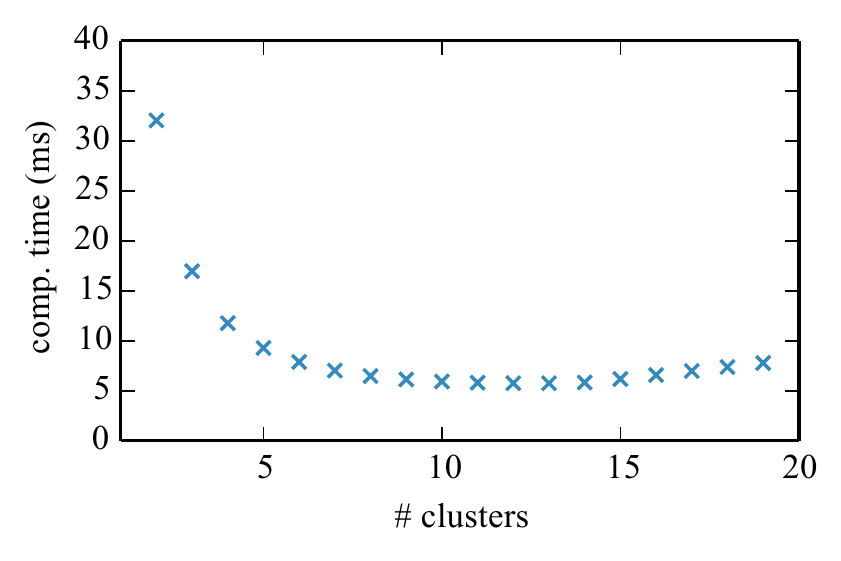}
\caption{Computational time for the adaptive softmax on the Bulgarian Europarl data, as a function of the number of clusters.
\label{fig:cluster_europarl_bg}}
\end{figure}

\paragraph{Finding the number of clusters.}
The only remaining free variable in our optimization is $J$, since the other parameters are then determined by the aforementioned optimizations.
We plot in Figure~\ref{fig:cluster_europarl_bg} the optimal computation time, as a function of the number of clusters $J$, according to our model.
We observe that a small number of clusters, between $10$ and $15$ gives the best computation time.
Moreover, we observe that using more than 5 clusters does not lead to significant gains in computational time (a couple of milliseconds at best).
In practice, we thus decide to use a small number of clusters (between 2 and 5), as it usually lead to slightly better perplexity, and we empirically determine the best speed/perplexity compromise on training data.
As shown later by our experiments, using a small number of clusters allows to obtain comparable perplexity as the exact softmax on large corpora.

\section{Experiments}
\label{sec:experiments}

This section provides a set of experiments aiming at analyzing the trade-off
between actual computation time and effectiveness of several strategies, in
particular the approach presented in the previous section.
First we describe our evaluation protocol, then we evaluate some of the properties of our model
and finally we compare it on standard benchmark against standard baselines.

\begin{table}
  \centering
  \begin{tabular}{lrr}
    \toprule
    & ppl\ \ & training time \\
    \midrule
    full softmax   & 144 & 83 min \\
    sampling       & 166 & 41 min \\
    HSM (freq)     & 166 & 34 min \\
    HSM (sim)      & 155 & 41 min \\
    D-softmax      & 195 & 53 min \\
    D-softmax [*]  & 147 & 54 min \\
    \midrule
    {\bf Ours}     & 147 & 30 min \\
    \bottomrule
\end{tabular}
\caption{Text8: perplexity and training time after 5 epochs. Our approach is significantly better than other published approximate strategies. We also show that improving the baseline D-softmax\,[*] as discussed in text improve the results, but is slower than our proposal. Note, approximate strategies are comparatively less interesting for small vocabularies such as in this case. 
\label{tab:text8}}
\end{table}

\begin{table*}[t]
\centering
\begin{tabular}{lc}
\toprule
Model & Test perplexity \\
\midrule
Interpolated Kneser-Ney 5-gram~\citep{chelba2013one}                 & 67.6 \\
Feedforward NN + D-Softmax~\citep{chen2015strategies}                & 91.2 \\
4-layer IRNN-512~\citep{le2015simple}                                & 69.4 \\
RNN-2048 + BlackOut sampling~\citep{ji2015blackout}                  & 68.3 \\
Sparse Non-negative Matrix Language Model~\citep{shazeer2015sparse}  & 52.9 \\
RNN-1024 + MaxEnt 9-gram~\citep{chelba2013one}                       & 51.3 \\
LSTM-2048-512~\citep{jozefowicz2016exploring}                        & 43.7 \\
2-layer LSTM-8192-1024 + CNN inputs~\citep{jozefowicz2016exploring}  & 30.0 \\
\midrule
\textbf{Ours} (LSTM-2048)                                            & 43.9 \\
\textbf{Ours} (2-layer LSTM-2048)                                    & 39.8 \\
\bottomrule
\end{tabular}
\caption{One Billion Word benchmark. Perplexity on the test set for single models. Our result is obtained after $5$ epochs.}
\label{tab:oneb}
\end{table*}

\begin{table*}[t]
\centering
{
\begin{tabular}{lrrrrrrrrrrrrrrrrrr}
\toprule
\multicolumn{1}{r}{Language:} &&
\multicolumn{2}{c}{bg} &&
\multicolumn{2}{c}{cs} &&
\multicolumn{2}{c}{da} &&
\multicolumn{2}{c}{de} &&
\multicolumn{2}{c}{el} &&
\multicolumn{2}{c}{es} \\ 
\multicolumn{1}{r}{$k$=} &&
\multicolumn{2}{c}{50k} &&
\multicolumn{2}{c}{83k} &&
\multicolumn{2}{c}{128k} &&
\multicolumn{2}{c}{143k} &&
\multicolumn{2}{c}{100k} &&
\multicolumn{2}{c}{87k} \\ 
Method          && ppl & $t$\ \  && ppl & $t$\ \ && ppl & $t$\ \ && ppl & $t$\ \ && ppl & $t$\ \ && ppl & $t$\ \ \\ 
\midrule
Full            && 37 & 58 && 62 & 132 && 37 & 713 && 42 & 802 && 38 & 383 && 30 & 536 \\ 
Sampling        && 40 & 29 && 70 & 53  && 40 & 247 && 45 & 262 && 41 & 144 && 32 & 217 \\ 
HSM (freq)      && 43 & 17 && 78 & 29  && 42 & 114 && 51 & 124 && 45 & 73  && 34 & 110 \\
HSM (sim)       && 39 & 25 && 68 & 43  && 38 & 150 && 43 & 154 && 39 & 98  && 30 & 147 \\
D-softmax       && 47 & 36 && 82 & 75  && 46 & 369 && 56 & 397 && 50 & 211 && 38 & 296 \\
D-softmax\,[*]  && 37 & 36 && 62 & 76  && 36 & 366 && 41 & 398 && 37 & 213 && 29 & 303 \\
{\bf Ours}      && 37 & 18 && 62 & 30  && 35 & 105 && 40 & 110 && 36 & 72  && 29 & 103 \\
\bottomrule
\end{tabular}
}
\caption{Europarl. Perplexity after 5 epochs for different languages as a function of time $t$ (minutes).
\label{tab:europar}}
\end{table*}

\paragraph{Datasets.}
We evaluate our method on standard datasets, and use the perplexity (ppl) as an evaluation metric, as the function of the training time or of the number of training data (epochs). 
The datasets have varying vocabulary sizes, in different languages, which allows us to better understand the strengths and weaknesses of the different approaches.
\begin{itemize}
\item
Text8\footnote{http://mattmahoney.net/dc/textdata} is a standard compression dataset containing a pre-processed version of 
the first $100$ million characters from Wikipedia in English.
It has been recently used for language modeling~\citep{mikolov2014learning} and has a vocabulary of $44$k words.
\item
Europarl\footnote{http://www.statmt.org/europarl/} is a machine translation corpus, containing 20 
languages~\citep{koehn2005europarl}. For most languages, there are
10M--60M tokens and the vocabulary is in between 44k and 250k words.
\item
One Billion Word~\footnote{https://code.google.com/archive/p/1-billion-word-language-modeling-benchmark/}
is a massive corpus introduced by \citet{chelba2013one}. It
contains $0.8$B tokens and a vocabulary comprising almost 800k words.
\end{itemize}

\paragraph{Implementation details.} We use an LSTM with one layer in all our experiments.
On Text8 and Europarl, the models have $d=512$ hidden units and are regularized with weight decay ($\lambda=10^{-6}$).
On the One Billion Word benchmark, we use $d=2048$ hidden units and no regularization.
The dimension of the input word embeddings is set to $256$, so that large models fit in GPU memory.
For the backpropagation through time, we unroll the models for 20 steps.
We use Adagrad~\citep{duchi2011adaptive}, with a step size of 0.1 and 5 epochs, and we clip the norm of the gradients to 1.
The batch size $B$ is set to 128, except on the Finnish portion of Europarl where $B$=64 due to memory constraints.
All the experiments were run on the same GPU with the Maxwell architecture.

\paragraph{Baselines.}
Our method is compared to: (1) the full softmax, (2)
the hierarchical softmax with frequency binning (HSM freq)
and similarity-based binning (HSM sim),
(3) importance sampling~\citep{bengio2003quick,bengio2008adaptive} and (4) the
differentiated softmax~\citep{chen2015strategies}.
For HSM, we tried different strategies for the binning.
We observe that using the square root function on the count before
computing the word bins is the most efficient for frequency binning.
For the similarity-based binning, we used the Brown clustering algorithm~\cite{brown1992class}
to determine the word classes.
For the negative sampling method, we used a number of samples equal to $20\%$
of the size of the vocabulary~\citep{chen2015strategies}. For the
differentiated softmax (D-softmax), we used the same partitions for the vocabulary as for
our approach.  We tried two version of the differentiated softmax. The first is the one described by 
\citet{chen2015strategies}, where each word cluster uses a disjoint subset of the hidden representation. 
We also present an improved version, referred to as D-softmax\,[*], which uses our choice to have the whole hidden representation mapped to the different word clusters using projection matrices of different sizes.

\begin{figure}[t]
\includegraphics[width=1\linewidth]{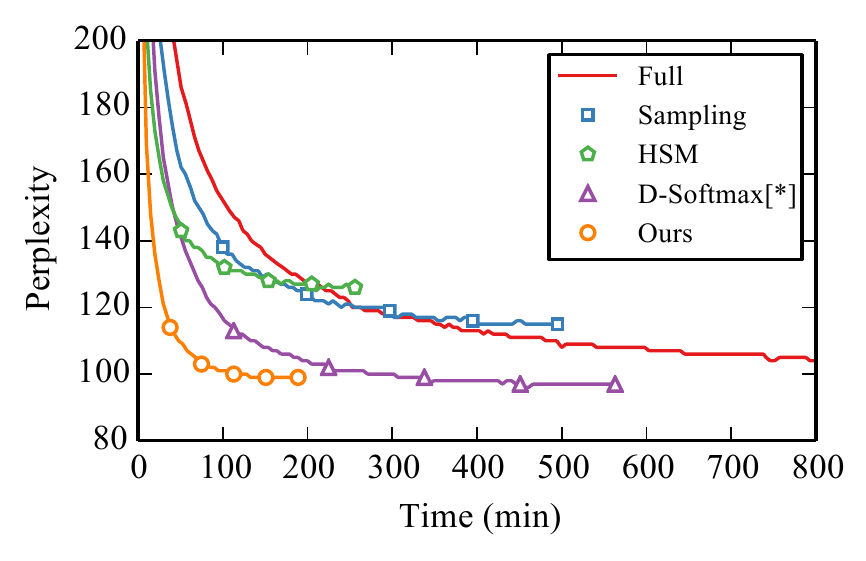}
\caption{Finnish Europarl: perplexity (on validation) as the function of time for our method and baselines.
We represent the result after each epoch by a point.
Our method favorably compares with all other approaches w.r.t. the tradeoff perplexity and training time.
Similar conclusions are drawn for the other languages. 
\label{fig:converge}}
\end{figure}

\paragraph{Comparison with the state of the art.}
Table~\ref{tab:text8} reports the results that we achieve on Text8. On this small vocabulary, approximate methods are comparatively less interesting. Our approach is the only one to approach the result of the full soft-max (below by 3 points of perplexity), while being the fastest. Our improved variant D-softmax\,[*] of the work by \citet{chen2015strategies} obtains similar results but is slower by a factor $\times 1.8$. 

On Europarl, we first present the convergence properties of our approach compared to other approximate strategies in Figure~\ref{fig:converge} show the perplexity (ppl) as a function of training time. Our approach significantly outperforms all competitors by a large margin. For reference, we also show the performance (D-softmax\,[*]) obtained by improving the D-softmax, to make it more comparable to our method. Our method is $2\times$ to $3\times$ faster than this improved competitor, which demonstrates how critical is our optimization strategy. Similar conclusions are drawn from Table~\ref{tab:europar} for other languages from the Europal corpus. 

Table~\ref{tab:oneb} gives the test perplexity on One Billion Word benchmark:
Our method achieves a perplexity of $43.9$ after five epochs, taking less than three days to train on a single GPU.
In comparison, only \citet{jozefowicz2016exploring} achieves a lower perplexity, but 
with a model $8\times$ bigger than ours and trained over $32$ GPUs during $3$ weeks.
We also note that for models of similar size, we achieve similar perplexity than the method introduced by \citet{jozefowicz2016exploring}.
As far as we know, ours the first method to achieve a perplexity lower than 50 on a single GPU.

\section{Conclusion}

In this paper, we have proposed a simple yet efficient approximation of the
softmax classifier. To our knowledge, it is the first speed optimizing approximation that
obtains performance on par with the exact model. This is achieved by explicitly
taking into account the computation time of matrix-multiplication on parallel systems and
combining it with a few important observations, namely keeping a short-list of
frequent words in the root node~\citep{schwenk2007continuous} and reducing the
capacity of rare words~\citep{chen2015strategies}.
In all our experiments on GPU, our method consistently
maintains a low perplexity while enjoying a speed-up going from $2\times$ to
$10\times$ compared to the exact model.
This type of speed-up allows to deal with extremely large corpora in reasonable
time and without the need of a large number of GPUs.
We believe our approach to be general enough to be applied to other parallel computing architectures and
other losses, as well as to other domains where the distributions of the class are unbalanced.

\label{sec:conclusion}

\subsection*{Acknowledgements}
The authors would like to thank Jeff Johnson for his help with GPU benchmarking as well as Tomas
Mikolov, Rob Fergus and Jeff Johnson for insightful discussions.

\small
\bibliography{softmax}
\bibliographystyle{icml2017}

\end{document}